\documentclass[a4paper,12pt,onecolumn]{article}

\usepackage{FAS}
\usepackage[latin1]{inputenc}
\usepackage{ngerman}
\usepackage{graphicx}
\usepackage{enumitem}
\usepackage{amsmath,amssymb,amsfonts}
\usepackage{algorithmic}
\usepackage{textcomp}
\usepackage{xcolor}
\usepackage{makecell}

\usepackage{multirow} % For table
\usepackage{tabularx} % For table
\usepackage{booktabs}
\usepackage{appendix}

\newcommand{\upperRomannumeral}[1]{\uppercase\expandafter{\romannumeral#1}}
\newcommand{\lowerromannumeral}[1]{\romannumeral#1\relax}

\makeatletter
\def\thanks#1{\protected@xdef\@thanks{\@thanks
        \protect\footnotetext{#1}}}
\makeatother

\begin{document}
%
% Titel:
\title{Beyond In-Distribution Performance: A Cross-Dataset Study of Trajectory Prediction Robustness}
\author{Yue Yao$^{1,2}$, Daniel Goehring$^2$, Joerg Reichardt$^1$
        % <-this % stops a space
%\thanks{}% <-this % stops a space
\thanks{The authors are with $^1$Continental Automotive GmbH (yue.yao@continental.com, joerg.reichardt@continental.com), $^2$Dahlem Center for Machine Learning and Robotics, Freie Universitaet Berlin (daniel.goehring@fu-berlin.de)}
% \thanks{Yue Yao is with Continental AG and Ph.D. candidate at Freie Universität Berlin.}%
% \thanks{Shengchao Yan is Ph.D. candidate at Universität Freiburg.}%
% \thanks{Daniel Goehring is Juniorprofessor for Autonomous Vehicles at the Dahlem Center for Machine Learning and Robotics of the Freie Universität Berlin}%
% \thanks{Joerg Reichardt is with Continental AG.}%
% \thanks{Wolfram Burgard is with the Department of Engineering, University of Technology Nuremberg, Germany.}
}
%
%
% Achtung: Datum soll nicht verwendet werden. Deshalb muss
%          leere Anweisung "\date{}" in n�chster Zeile stehen!
\date{}

% erzeuge Titel
\maketitle 

\begin{abstract}
 % The ability to handle Out-of-Distribution (OoD) samples is an essential metric for evaluating the robustness of trajectory prediction models.  However, the development and ranking of state-of-the-art (SotA) models are driven by their In-Distribution (ID) performance on individual competition datasets. This work extensively evaluates the ID performance and OoD generalization of three SotA models across two large-scale motion datasets, Argoverse 2 (A2) and Waymo Open Motion (WO). We highlight the improved model generalization using polynomial data representation and homogeneous data augmentation by comparing our proposed model against benchmark methods. However, our experiments demonstrate an unexpected generalization trend under a reversed training and testing setup: models trained on the larger WO dataset exhibited diminished robustness when tested on the smaller A2 dataset. We hypothesize that this is influenced by the complexity of prediction tasks and dataset noise characteristics, highlighting the complex interplay between dataset properties and model design. Our findings emphasize the importance of OoD testing as an essential performance metric alongside traditional in-distribution evaluation.

% Alternative which duplicates less of the introduction
We study the Out-of-Distribution (OoD) generalization ability of three SotA trajectory prediction models with comparable In-Distribution (ID) performance but different model designs. We investigate the influence of inductive bias, size of training data and data augmentation strategy by training the models on Argoverse 2 (A2) and testing on Waymo Open Motion (WO) and vice versa.
We find that the smallest model with highest inductive bias exhibits the best OoD generalization across different augmentation strategies when trained on the smaller A2 dataset and tested on the large WO dataset. In the converse setting, training all models on the larger WO dataset and testing on the smaller A2 dataset, we find that all models generalize poorly, even though the model with the highest inductive bias still exhibits the best generalization ability. We discuss possible reasons for this surprising finding and draw conclusions about the design and test of trajectory prediction models and benchmarks.
\end{abstract}


%%%%%%%%%%%%%%%%%%%%%%%%%%%%%%%%%%%%%%%%%%%%%%%%%%%%%%%%%%
\section{Introduction}
\label{secs: introduction}
The robustness of trajectory prediction is essential for practical applications in autonomous driving. The advancement of trajectory prediction models is catalyzed through public motion datasets and associated competitions, such as Argoverse 2  (A2) \cite{wilson_argoverse2_2021}, and Waymo Open Motion (WO) \cite{ettinger_waymo_2021}. These competitions establish standardized metrics and test protocols and score predictions on test data that is withheld from all competitors and hosted on protected evaluation servers only. This is intended to objectively compare the generalization ability of models to unseen data. 

However, these withheld test examples still share similarities with the training samples, such as sensor setup, map representation, post-processing, geographic, and scenario selection biases employed during dataset creation. Consequently, the test scores reported in each competition are examples of \emph{In-Distribution} (ID) testing. To effectively evaluate model generalization, it is essential to test models on truly \emph{Out-of-Distribution} (OoD) test samples, such as those from different motion datasets.

% The differences in data collection processes and sensor platforms across motion datasets from different origins provide an opportunity for \emph{Out-of-Distribution} (OoD) testing. This allows for the evaluation of model generalization on truly independent data samples.
% However, cross-dataset testing requires working around inconsistencies in data formats and prediction tasks between datasets, such as different prediction horizons. We address this in our prior work by introducing a homogenizing protocol and standardize data formats and prediction tasks across two large-scale datasets \cite{yao_improving_2024}, Argoverse 2 (A2) and Waymo Open Motion (WO).

We investigate model generalization across two large-scale motion datasets \cite{yao_improving_2024}: Argoverse 2 (A2) and Waymo Open Motion (WO). The WO dataset, with 576k scenarios, is more than twice the size of A2, which contains 250k scenarios.
As model generalization ability is best illustrated by limiting the amount of training data \cite{hestness_deep_2017}, we first train on the smaller dataset A2 and test on OoD samples from the larger dataset WO \cite{yao_improving_2024}. Our results demonstrate that model generalization improves through two key factors: introducing inductive bias via polynomial representation (cf. Section \ref{sec: model data representation}) and implementing proper data augmentation (cf. Section \ref{sec: model data augmentation}). 

This naturally raises the question: how does model generalization change when the experimental setup is reversed, i.e., when models are trained on the larger WO dataset and tested on OoD samples from the smaller A2 dataset? Given the increased amount of training data, one would hypothesize that all models will exhibit improved generalization in OoD tests. In this study, we investigate model generalization under this reversed experimental setting to assess whether the hypothesized improvements in robustness are realized. %  Our key contributions are summarized as follows:

Our paper is organized as follows: We first detail the two design choices influencing model generalization -- data representation and data augmentation -- identified in our prior work \cite{yao_improving_2024}. From the body of related work, we introduce the two benchmark models along with our proposed model and detail their model designs. We then revisit our dataset homogenization protocol for OoD testing and outline the new experimental settings for training on the WO dataset. Subsequently, we present ID and OoD testing results for all models and analyze the new OoD results with respect to data representation, augmentation strategies and compare them to our expectations. Finally, we conclude our work by arguing for establishing OoD testing as an equally important performance metric alongside traditional ID evaluation.

\section{Related Work and Benchmark Models}
% \subsection{Data Noise}
% The noise in training and test data significantly influences model performance. Studies in the computer vision domain have demonstrate that deep neural networks can easily overfit to noisy training samples due to their high capacity \cite{song_learning_2022}. The noise characteristics in motion data are associated with multiple factors, such as sensor setup and post-processing. However, the details of noise are not reported in the motion dataset, making it challenging to study and address data noise effectively. 

% Resent studies emphasize the presence of outliers in multiple large-scale motion datasets \cite{yao_empirical_2023, hu_processing_2022}. In our recent work \cite{yao_empirical_2023}, we reverse-engineered the noise characteristics using empirical Bayes analysis across multiple large-scale motion datasets. The impact of data noise on the robustness of trajectory prediction models remains underexplored, presenting us as the primary motivation for this work.

\subsection{Data Representation}
\label{sec: model data representation}

The structure and rules of individual datasets and accompanying competitions heavily influence the design choices of trajectory prediction models in the literature. Deep learning-based models often directly ingest data in the format the dataset is stored in.
Recent studies employ \emph{sequence-based representation}, e.g., sequences of data points, as the model's input and output \cite{cheng_forecast_2023, zhou_query_2023}. This approach effectively captures diverse information, including agent trajectories and road geometries, while aligning with dataset measurement formats. However, it has notable drawbacks: the representation introduces high redundancy, and its computational cost scales with the length and temporal/spatial resolution of the trajectories and road geometries. Furthermore, the high flexibility of this approach can capture measurement noise and outliers, potentially leading to physically infeasible predictions. %Additionally, the computational requirement of this approach scales with the length and temporal/spatial resolution of trajectories and road geometries.

Some previous works explored the possibility of using \emph{polynomial representations} for predictions \cite{buhet_plop_2020, su_temporally_2021}. Su et al.\ \cite{ su_temporally_2021} highlight the temporal continuity of this representation, i.e., the ability to provide arbitrary temporal resolution. Reichardt \cite{reichardt_trajectories_2022} argues for using polynomial representations to integrate trajectory tracking and prediction into a filtering problem. Polynomial representations restrict the kind of trajectories that can be represented and introduce bias into prediction systems. This limited flexibility is generally associated with greater computational efficiency, smaller model capacity, and hence better generalization. 

In our previous study, we proposed our model EP with polynomial inputs and outputs as key innovations \cite{yao_improving_2024}. Our EP model demonstrates improved robustness against OoD samples compared to sequence-based SotA models. 

\subsection{Data Augmentation}
\label{sec: model data augmentation}
In competition settings, one or more agents in a scenario are typically designated as \emph{focal agents}, and predictions are scored exclusively for these agents. However, training the model exclusively with the focal agent's behavior fails to exploit all available data. To address this, predicting the future motion of non-focal agents is a typical augmentation strategy for training. As there are many more non-focal agents than focal agents, another important design decision involves determining how to balance the contributions of focal and non-focal agent data.

We select two open-sourced and thoroughly documented SotA models: Forecast-MAE \cite{cheng_forecast_2023} (FMAE) and QCNet \cite{zhou_query_2023}, with nearly 1900k and 7600k parameters, respectively. As summarized in Table \ref{tab: model difference in non-focal}, both models employ sequence-based representation but employ different strategies in dealing with non-focal agents:

\begin{itemize}[leftmargin=*]
    \item[] \textbf{Heterogeneous Augmentation}: FMAE follows the prediction competition protocol and prioritizes focal agent prediction. Thus, agent history and map information are computed within the \emph{focal agent's} coordinate frame. Compared to the \emph{multi-modal} prediction of the focal agent, FMAE only outputs \emph{uni-modal} prediction for non-focal agents. The error of focal agent predictions is weighed higher than for non-focal agents in loss function.
    \item[] \textbf{Homogeneous Augmentation}: QCNet does not focus on the selected focal agent and proposes a more generalized approach. It encodes the information of agents and map elements in each agent's \emph{individual} coordinate frame. It outputs \emph{multi-modal} predictions for focal and non-focal agents alike. This approach ensures a consistent prediction task for all agents and enhanced model generalization. The prediction error of focal and non-focal agent predictions is weighed equally in the loss functions.
\end{itemize}

\begin{figure}[thb]
\centering
\includegraphics[width=3.2in]{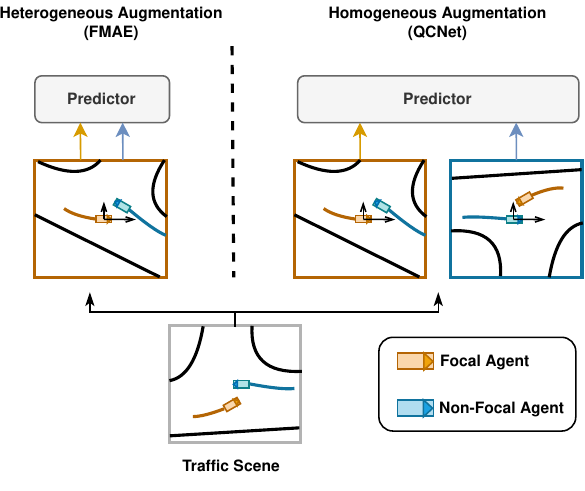}
\caption{The two augmentation strategies for non-focal agent data employed in benchmark models. \textbf{Left}: FMAE employs heterogeneous augmentation, representing information in the focal agent's coordinate frame and only making uni-modal predictions for non-focal agents. \textbf{Right}: QCNet employs homogeneous augmentation, encoding information in each agent's individual coordinate frame and making multi-modal predictions for both focal and non-focal agents alike.}
\label{fig:augmentation_strategy}
\vspace{-0.5em}
\end{figure}

\noindent Figure \ref{fig:augmentation_strategy} illustrates the augmentation strategies of our benchmark models FMAE and QCNet.

\begin{table}[!b]
\caption{Model variants under study based on augmentation strategies and data representations.}
\vspace{+0.5em}
\centering
\begin{tabularx}{5.5in}{l >{\centering\arraybackslash}X |>{\centering\arraybackslash}X }
\Xhline{3\arrayrulewidth}
augmentation & \multicolumn{2}{!{\vrule width 1pt}c}{input and output representation} \\
\cline{2-3}
strategy&\multicolumn{1}{!{\vrule width 1pt}c|}{sequence-based} &  polynomial-based (ours) \\
\Xhline{3\arrayrulewidth}
heterogeneous & \multicolumn{1}{!{\vrule width 1pt}c|}{\multirow{1}{*}{FMAE \cite{cheng_forecast_2023}}}&\multirow{1}{*}{EP-F}\\
%+&\multicolumn{1}{!{\vrule width 1pt}c|}{}   \\
\cline{1-3}
homogeneous & \multicolumn{1}{!{\vrule width 1pt}c|}{\multirow{1}{*}{QCNet \cite{zhou_query_2023}}} & \multirow{1}{*}{EP-Q} \\
%+&  \multicolumn{1}{!{\vrule width 1pt}c|}{}  &  \\
\cline{1-3}
\multirow{2}{*}{w/o augmentation} & \multicolumn{1}{!{\vrule width 1pt}c|}{FMAE-noAug} & \multirow{2}{*}{EP-noAug} \\
&\multicolumn{1}{!{\vrule width 1pt}c|}{QCNet-noAug}& \\
% &  \multicolumn{1}{!{\vrule width 1pt}c|}{QCNet-}  &  \\
%&\multicolumn{1}{!{\vrule width 1pt}c|}{}   &  \\
\Xhline{3\arrayrulewidth}
\end{tabularx}
\label{tab: model difference in non-focal}
%\vspace{-1.em}
\end{table}

Following our previous study \cite{yao_improving_2024}, we introduce the third augmentation strategy for comparison: \textbf{no augmentation}. This can simply be achieved by removing the loss of non-focal agents in each model, thus limiting the model to learn from focal agent behavior only. We denote the benchmark models without augmentation as ``FMAE-noAug'' and ``QCNet-noAug''.

To evaluate the impact of data representation independently of augmentation strategies, we implement all three augmentation modes for our proposed model, EP. These include: ``EP-noAug'' (no augmentation), ``EP-F'' (heterogeneous augmentation), and ``EP-Q'' (homogeneous augmentation).

%%%%%%%%%%%%%%%%%%%%%%%%%%%%%%%%%%%%%%%%%%%%%%%%%%%%%%%%%%
\section{Dataset Homogenization}
% The differences in the data collection processes and sensor
% platforms between motion datasets of different origins present
% us with the opportunity to perform OoD testing on truly
% independent data samples. However, this also comes with the
% challenges of working around inconsistencies in data formats
% and prediction tasks between datasets.

To work around inconsistencies in data formats
and prediction tasks between datasets, we inherit the homogenization protocol from the one proposed in prior work \cite{yao_improving_2024} with a different \emph{prediction horizon}. This adopts the following settings for both datasets and details are summarized in Table \ref{tab: dataset difference}:
\begin{itemize}[leftmargin=*]
    \item History Length: We set the history length to 5 seconds (50 steps) as in A2.
    \item Prediction Horizon:
    All models are trained with 4.1s prediction due to the limited recording length of training data in WO.
    \item Map Information: We exclude boundary information and the label of lane segments in junctions due to the information's absence in WO. Only lane segments and crosswalks are considered map elements in homogenized data due to their availability in both datasets.
    \item Focal Agent: We take the same focal agent as labeled in A2. From WO, only the first fully observed, non-ego agent in the list of focal agents is chosen. As the list of focal agents is unordered, this corresponds to randomly sampling a single fully observed focal agent.
\end{itemize}

\begin{table*}[!th]
\caption{Datasets comparison and homogenization protocol for OoD testing.}
\centering
\begin{tabularx}{6.0in}{l l >{\centering\arraybackslash}X >{\centering\arraybackslash}X >{\centering\arraybackslash}X}
\Xhline{3\arrayrulewidth}
\multicolumn{2}{c}{\multirow{2}{*}{} } & \multirow{2}{*}{A2 ~\cite{wilson_argoverse2_2021}} & \multirow{2}{*}{WO~\cite{ettinger_waymo_2021}} & Homogenized Dataset\\
%&&&& train \:\vline\: test\\
\Xhline{3\arrayrulewidth}
\multicolumn{2}{l}{scenario length}& 11s & 9.1s & 9.1s \\
\hline
\multicolumn{2}{l}{sampling rate}& 10 Hz & 10 Hz & 10 Hz\\
\Xhline{3\arrayrulewidth}
\multicolumn{2}{l}{history length}& 5s & 1.1s & 5s \\
\hline
\multicolumn{2}{l}{\multirow{1}{*}{prediction horizon}} & \multirow{1}{*}{6s}  & \multirow{1}{*}{8s } & 4.1s \\
%& &&&(60 steps) \vline\:(41 steps)  \\
\hline
\multirow{5}{*}{\thead{map\\information}}  & lane center & yes & \multirow{1}{*}{yes} & \multirow{1}{*}{yes} \\
\cline{2-5}
& \multirow{3}{*}{lane boundary} & \multirow{3}{*}{yes} & not for junction lanes& \multirow{3}{*}{no}\\
\cline{2-5}
& junction lanes labeled & yes& no& no \\
%\cline{2-5}
%& other map elements & \multirow{2}{*}{no} & \multirow{2}{*}{yes} & \multirow{2}{*}{no}\\
%& (e.g. traffic lights) &&& \\
\hline
\multirow{4}{*}{\thead{prediction\\target}}  & \# focal agents & 1 & up to 8 & 1\\
\cline{2-5}
& ego included & no & yes & no\\
\cline{2-5}
& \multirow{2}{*}{fully observed}& \multirow{2}{*}{yes} & not guaranteed & \multirow{2}{*}{yes} \\
\Xhline{3\arrayrulewidth}
\end{tabularx}
\label{tab: dataset difference}
\end{table*}

\noindent Based on the focal agent's selection criterion above, we have
472 235 training samples and 42 465 validation samples from WO.
For the A2 dataset, we have 199908 and 24 988 valid samples from training and validation split, respectively.

%%%%%%%%%%%%%%%%%%%%%%%%%%%%%%%%%%%%%%%%%%%%%%%%%%%%%%%%%%
\section{Experimental Results}
\subsection{Experiment Setup and Metrics}
\label{secs: experiment setup}
All models are trained on WO from scratch using the original hyperparameters. The WO dataset features a more complex map structure, with more lanes and map points than A2. This leads to ``out of memory'' issue on our GPU cluster with 192 GB of RAM when training QCNet with its default settings. To fit QCNet on the available training hardware, we had to limit the scenario complexity for QCNet to process up to 50 agents and 80 map elements closest to the ego vehicle per scene. %The WO validation set is used as ID test samples, while the A2 validation set serves as OoD test samples. %To mitigate this, we limit QCNet to process up to 50 agents and 80 map elements closest to the ego vehicle per scene. The WO validation set is used as ID test samples, while the A2 validation set serves as OoD test samples.

For ID testing, we use the official benchmark metrics, including minimum Average Displacement Error ($minADE_{K}$) and minimum Final Displacement Error
($minFDE_{K}$), for evaluation. The metric $minADE_{K}$  calculates the average Euclidean distance between the ground-truth trajectory and the best of K predicted trajectories across all future time steps, while the $minFDE_{K}$ measures the final-step error, focusing on long-term accuracy.

For OoD testing, we also propose $\Delta minADE_{K}$ and $\Delta minFDE_{K}$ as the difference of displacement error between ID and OoD testing to measure model robustness. In accordance with standard practice, $K$ is selected as 1 and 6.

\subsection{ID Testing Results}
\label{secs: ID results}
We present our new ID results in the upper section of Table \ref{tab: in-distribution result combined} and reference the ID results from our previous study in the lower section for comparison. In the upper section, all models are trained on the homogenized WO training split and tested on the homogenized WO validation split. As in prior work, the QCNet results serve as the reference for relative comparisons, i.e., as $100\%$. 

As highlighted in \cite{feng_unitraj_2025}, the WO dataset features a significantly larger volume of data and higher variance, presenting challenges for small models for training and inference. However, with the much smaller model size (345k), both EP-F and EP-Q outperform FMAE and achieve comparable ID performance to QCNet in $minADE_1$ and $minFDE_1$ for the new ID results. Despite limiting the scenario complexity for QCNet, it still achieves the best ID performance in 3 out of 4 metrics.

Also, note that all models perform better on WO than on A2 in absolute error under all augmentation settings in ID tests, except the $minFDE_6$ of FMAE increase from $0.792$m to $0.829$m. The better ID performance on WO can be attributed to the larger size of the WO dataset, the relative simplicity of WO scenarios, or a combination of both factors. %This is either a result of the WO dataset being larger or the scenarios in WO being easier to predict or a combination of both factors.

% Compared to our previous study, we observe two notable differences: 

% \begin{itemize}[]
%     \item[] \textbf{Lower Prediction Errors}: All models exhibit reduced prediction errors compared to their ID results on A2, except for FMAE's $minFDE_6$ which shows a higher error (0.829m vs. 0.792m).
%     \item[] \textbf{Reversed Ranking of EP-Q}: In the previous study (lower section), EP-Q showed the worst ID performance compared to FMAE and EP-F when trained and tested on A2. However, in the new results (upper section), EP-Q outperforms both FMAE and EP-F, achieving the lowest prediction errors across all metrics.
% \end{itemize}

% We hypothesize that both observations are attributed to the pronounced focal agent selection bias present in A2 \cite{wilson_argoverse2_2021}. The behavior of focal agents at specifically curated timesteps is more challenging to predict, benefiting models with heterogeneous augmentation and enhancing their ID performance on A2. Conversely, the homogenized WO dataset with modified history length and prediction horizon exhibits less bias in focal agent behavior, resulting in comparatively worse ID performance for FMAE and EP-F on WO.

\begin{table*}[tbh]
%\vspace{+5pt}
\caption{Result of In-Distribution (ID) testing of QCNet, FMAE, EP-Q and EP-F evaluated with a 4.1-second prediction horizon. For clarity, the results of models without augmentation are not reported. \textbf{Upper Section}: ID result of models trained on the homogenized WO training set and tested on the homogenized WO validation set. \textbf{Lower Section}: Referenced ID result of models trained on the homogenized A2 training set and tested on the homogenized A2 validation set \cite{yao_improving_2024}. }

\vspace{+5pt}
\centering

    \begin{tabularx}{\textwidth}{>{\centering\arraybackslash}X c | >{\centering\arraybackslash}X >{\centering\arraybackslash}X >{\centering\arraybackslash}X >{\centering\arraybackslash}X }
    \toprule
    \multirow{2}{*}{\shortstack{Training}} & \multirow{2}{*}{\shortstack{Models}} & \multicolumn{1}{c}{minADE$_{1}$} & \multicolumn{1}{c}{minFDE$_{1}$} & \multicolumn{1}{c}{minADE$_{6}$} & \multicolumn{1}{c}{minFDE$_{6}$} \\
        &&[m]&[m]&[m]&[m] \\
        \midrule
        \multirow{8}{*}{\shortstack{\thead{WO}}} & \multirow{2}{*}{QCNet \cite{zhou_query_2023}} &0.820 & 2.171 & 0.344& 0.696 \\
        & &(100.0\%)&(100.0\%)&(100.0\%)&(100.0\%)\\
        \cmidrule(lr){3-6} 
        & \multirow{2}{*}{\thead{FMAE \cite{cheng_forecast_2023} \\ w/o pre-train }} &0.889 & 2.318& 0.374 & 0.829 \\
        &&(108.4\%)&(106.8\%)&(108.7\%)&(119.1\%)\\
        \cmidrule(lr){3-6} 
        & \multirow{2}{*}{EP-Q (ours) \cite{yao_improving_2024}} &0.821 & 2.155 & 0.359 & 0.802\\
        &&(100.1\%)&(99.3\%)&(104.4\%)&(115.2\%)\\
        \cmidrule(lr){3-6} 
        & \multirow{2}{*}{EP-F (ours) \cite{yao_improving_2024}} & 0.831 & 2.171  & 0.370 & 0.825\\ 
        &&(101.3\%) & (100.0\%) & (107.6\%) & (118.5\%)\\
        \bottomrule

        \multirow{8}{*}{\shortstack{\thead{A2}}} & \multirow{2}{*}{QCNet \cite{zhou_query_2023}} &0.902 & 2.301 & 0.400& 0.701 \\
        &&(100.0\%)&(100.0\%)&(100.0\%)&(100.0\%)\\
        \cmidrule(lr){3-6} 
        & \multirow{2}{*}{\thead{FMAE \cite{cheng_forecast_2023} \\ w/o pre-train }} &0.989 & 2.459& 0.423 & 0.792 \\
        &&(109.4\%)&(106.9\%)&(105.8\%)&(113.0\%)\\
        \cmidrule(lr){3-6} 
        & \multirow{2}{*}{EP-Q (ours) \cite{yao_improving_2024}} & 1.161 & 2.830 & 0.491 & 0.922\\
        &&(128.7\%)&(123.0\%)&(122.8\%)&(131.5\%)\\
        \cmidrule(lr){3-6} 
        & \multirow{2}{*}{EP-F (ours) \cite{yao_improving_2024}} & 1.082 & 2.555  & 0.476 & 0.865\\ 
        &&(120.0\%) & (111.0\%) & (119.0\%) & (123.4\%)\\
        \bottomrule
    \end{tabularx}
    \label{tab: in-distribution result combined}
    \vspace{-5pt}
\end{table*}

\begin{figure}[!t]
%\vspace{+5pt}
\centering
\includegraphics[width=6.2in]{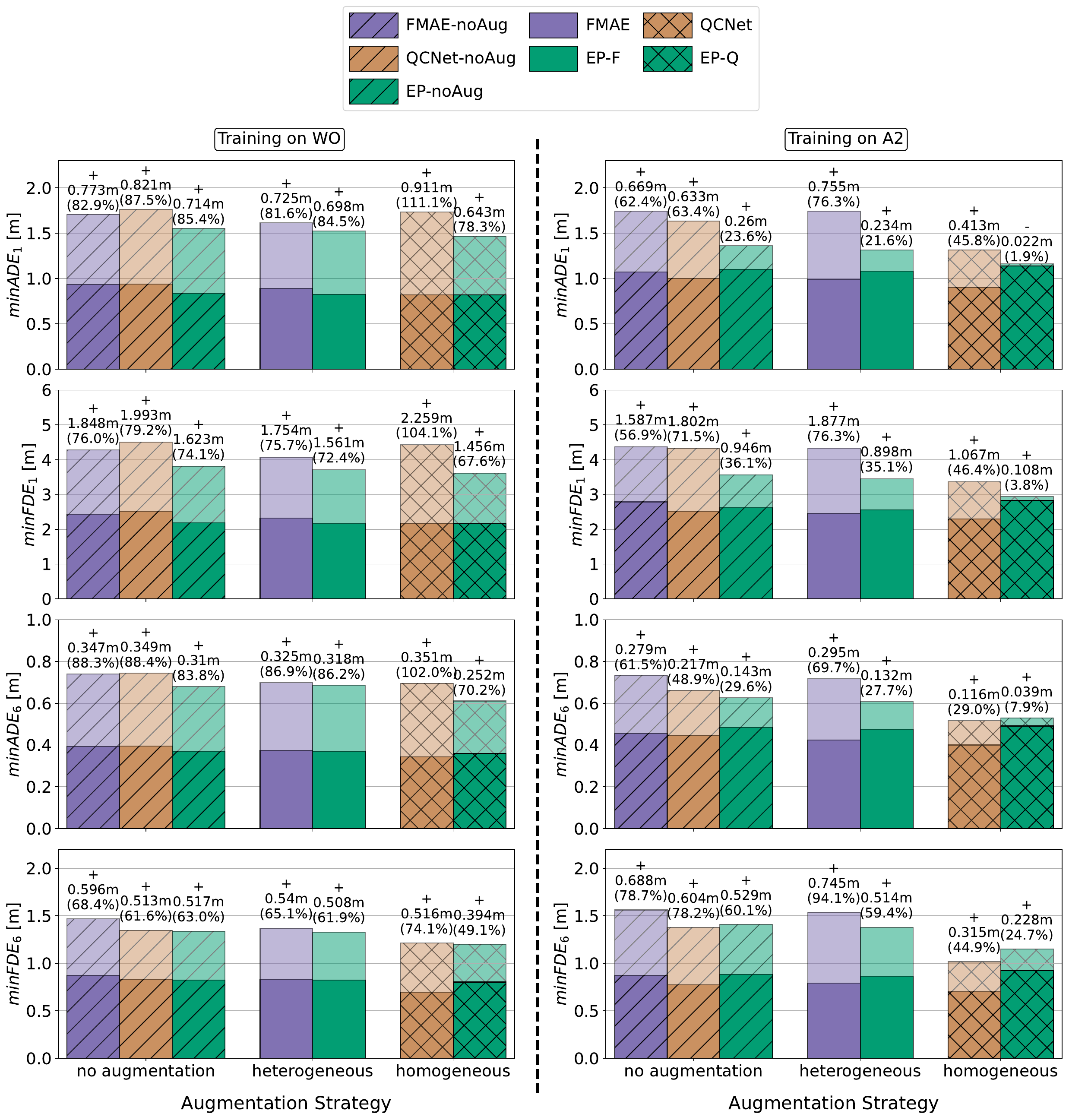}
\caption{The OoD testing results of FMAE, QCNet, EP and their variants. We indicate the \emph{absolute} and \emph{relative} difference in displacement error between ID  and OoD results. The \emph{solid} bars represent ID results reported in Table \ref{tab: in-distribution result combined}, while the \emph{transparent} bars indicate the increase in displacement error during OoD testing. \textbf{Left}: Models trained on the homogenized WO training set and tested on the homogenized WO (ID) and A2 (OoD) validation sets. \textbf{Right}: Models trained on the homogenized A2 training set and tested on the homogenized A2 (ID) and WO (OoD) validation sets \cite{yao_improving_2024}.}
\label{fig:OoD_results_combined}
\vspace{0.0em}
\end{figure}

% \begin{figure}[!t]
% %\vspace{+5pt}
% \centering
% \includegraphics[width=6.2in]{images/out of dist results combined reversed.pdf}
% \caption{The OoD testing results of FMAE, QCNet, EP and their variants. We indicate the \emph{absolute} and \emph{relative} difference in displacement error between ID  and OoD results. The \emph{solid} bars represent ID results reported in Table \ref{tab: in-distribution result combined}, while the \emph{transparent} bars indicate the increase in displacement error during OoD testing. \textbf{Left}: Models trained on the homogenized A2 training set and tested on the homogenized A2 (ID) and WO (OoD) validation sets \cite{yao_improving_2024}. \textbf{Right}: Models trained on the homogenized WO training set and tested on the homogenized WO (ID) and A2 (OoD) validation sets.}
% \label{fig:OoD_results_combined}
% \vspace{0.0em}
% \end{figure}

\subsection{OoD Testing Results}
Figure \ref{fig:OoD_results_combined} compares our new OoD results (left) with our previous OoD results (right). In both cases, we evaluate model robustness and generalization using the absolute and relative increases in prediction error, e.g. $\Delta minADE_{K}$ and $\Delta minFDE_{K}$.
% We reference our previous OoD results as the right part and visualize our new OoD results in the left part of Figure . The relative and absolute increase in the prediction metric, i.e.  will serve as our measure of robustness and generalization ability. 
We will present the OoD results from three perspectives: (\lowerromannumeral{1}) augmentation strategy, (\lowerromannumeral{2}) data representation, and (\lowerromannumeral{3}) comparison with our expectations in Section \ref{secs: introduction}.

\subsubsection{Augmentation Strategy}
Our previous study (right side of Figure \ref{fig:OoD_results_combined}, originally) showed that heterogeneous augmentation yields only marginal improvements in robustness for OoD testing. In our new experiments (left side of Figure \ref{fig:OoD_results_combined}, originally), this trend continues for both EP-F and FMAE, indicating consistent results with respect to heterogeneous augmentation.
% Our previous study (right) showcases that heterogeneous augmentation only yields marginal improvement in model robustness in OoD testing. Our new OoD results align with our prior findings and demonstrate the consistent result for EP-F and FMAE. % Compared to variants without augmentation, models employing heterogeneous augmentation, such as FMAE and EP-F, only yields marginal improvement on model robustness in OoD testing. This aligns with our prior findings.

Based on our prior findings, we expect models employing homogeneous augmentation to exhibit a notable improvement in generalization. The results for EP-Q align with this expectation, as $\Delta minADE_6$ decreases from $+0.310$m $(83.8\%)$ to $+0.252$m
$(70.2\%)$ compared to EP-noAug. However, QCNet shows contrasting results with worse robustness compared to QCNet-noAug, with $\Delta minADE_6$ increasing from $+0.349$m
$(88.4\%)$ to $+0.351$m $(102.0\%)$. The performance of QCNet can be attributed to the limited scenario complexity, as discussed in Section \ref{secs: experiment setup}.

\subsubsection{Data Representation}
The advantage of using polynomial representation for model generalization remains evident for EP-Q, with a significant decrease in $\Delta minFDE_6$ from $+0.516$m
$(74.1\%)$ to $+0.394$m $(49.1\%)$ compared to QCNet. Moreover, EP-Q achieves the lowest prediction error in the new OoD testing across all benchmarks.

In contrast, while EP-F and EP-noAug still demonstrate improvement in robustness compared to sequence-based models, the gains are only marginal. This is reflected in the $\Delta minFDE_6$ between EP-F ($+0.508$m $(61.9\%)$) and FMAE ($+0.540$m $(65.1\%)$).
% However, though EP-F and FMAE still exhibit robustness improvement, evidenced by the $\Delta minFDE_6$ between EP-F and FMAE ($+0.508$m $(61.9\%)$ vs. $+0.540$m
% $(65.1\%)$), the improvement is marginal.
% For ``no augmentation'' and ``heterogeneous augmentation'' strategies, EP models only show marginally improved robustness compared to benchmarks using sequence-based data, as evidenced by the $\Delta minFDE_6$ between EP-F and FMAE ($+0.508$m $(61.9\%)$ vs. $+0.540$m
% $(65.1\%)$). In contrast, EP-Q still significantly reduces the prediction error increase against OoD test samples compared to QCNet, e.g. $+0.394$m $(49.1\%)$ vs. $+0.516$m
% $(74.1\%)$ in $\Delta minFDE_6$, and achieves the lowest prediction error in OoD testing across all benchmarks. %Furthermore, we observe a consistent reversal in the rankings of EP-Q and QCNet in ID and OoD evaluations based on $minADE_6$, mirroring the trends identified with $minADE_1$ in our prior work.

\subsubsection{Comparison with Expectations}
Comparing the new and previous OoD results, visualized in the left and right parts of Figure \ref{fig:OoD_results_combined}, respectively, our empirical observations diverge significantly from our expectations outlined in Section \ref{secs: introduction}. For instance, FMAE does not exhibit any improvement in robustness, as indicated by the nearly identical $\Delta minADE_1$ across both settings. Moreover, QCNet and EP variants exhibit significantly poorer generalization under the new settings when trained on WO. Notably, EP-Q's $\Delta minADE_1$ increases from  $-0.022$m $(1.9\%)$ in previous OoD results to $+0.643$m $(78.3\%)$ under the new settings. Drawing from our experimental observations and previous work on dataset characteristics, we provide two potential reasons: 

\begin{itemize}[]
    \item[] \textbf{Complexity of Prediction Task}: Creators of both datasets mined their recordings to identify challenging scenarios and focal agents for prediction \cite{wilson_argoverse2_2021, ettinger_waymo_2021}. Agent's motion becomes particularly difficult to predict when they accelerate, brake, or turn due to interactions with other agents and map - scenarios that simple models such as constant velocity cannot easily capture. While we maintain the history length of A2 for homogenized datasets, we increase the history length of WO from 1.1 second to 5 seconds. This adjustment potentially moves the challenging behaviors into the historical data, effectively making the prediction task less complex for WO. As a result, models trained on WO with these less challenging prediction tasks tend to generalize poorly when tested on A2, which features more demanding prediction scenarios. We present our experimental results to support this hypothesis in Appendix A.
    \item[] \textbf{Noise Level in Datasets}: Based on our previous study of the noise characteristics across both datasets \cite{yao_empirical_2023}, WO features lower noise levels compared to A2. Homogeneous augmentation and polynomial representation can be effective for training on a noisy dataset, such as A2, enhancing robustness with cleaner OoD test data (WO). Conversely, when the test data exhibits higher noise levels than the training data, both methods have only a minimal improvement in model robustness.
\end{itemize}

To validate these hypotheses, further research is required, such as including controlled experiments with known noise levels or identifying focal agent behavior. However, both hypotheses are closely tied to dataset properties, making it challenging to isolate and control variables for a definitive explanation. We leave this as an open question for future investigation.

%Based on our previous study of the noise characteristics across both datasets \cite{yao_empirical_2023}, WO features lower noise levels compared to A2. This indicates that homogeneous augmentation and polynomial representation are effective for training on a noisy dataset, such as A2, enhancing robustness with cleaner OoD test data (WO). Conversely, when the test data exhibits higher noise levels compared to the training data, both methods have only a minimal improvement on model robustness.
%%%%%%%%%%%%%%%%%%%%%%%%%%%%%%%%%%%%%%%%%%%%%%%%%%%%%%%%%%
\section{Conclusion and Future Directions}    
Our work emphasizes the importance of evaluating trajectory prediction models beyond In-Distribution (ID) performance, with a focus on Out-of-Distribution (OoD) robustness for autonomous driving. We systematically investigate the ID and OoD performance of three state-of-the-art prediction models and demonstrate the improved model robustness of our proposed EP-Q model with polynomial data representation and homogeneous data augmentation.

However, our findings show that model robustness is influenced by factors beyond model architecture and design choices. Contrary to our initial hypothesis, models trained on the larger Waymo Open Motion dataset showed reduced robustness when tested on the smaller Argoverse 2 dataset. This surprising result highlights the complex relationship between dataset properties, model design choices, and generalization performance. Our work outlines two potential factors for future investigation: the complexity of the prediction task and dataset noise levels. Our results outline the importance of OoD testing for prediction models and emphasize that improving OoD robustness requires a deeper understanding of both model design and dataset properties, rather than simply increasing training data volume.

%Analyzing potential factors is challenging without the support of the dataset creator, such as providing measurement noise. This work represents an initial step toward a trajectory prediction model that is capable of generalizing to different datasets and is robust under changes in sensor setup, scenario selection strategy or post-processing of training data.
% Recent SotA trajectory prediction models excel in In-Distribution (ID) performance but often struggle with Out-of-Distribution (OoD) scenarios, where rankings can reverse. Our proposed OoD testing protocol offers a new approach for evaluating model robustness, demonstrating the advantages of homogeneous augmentation and polynomial representations in our EP model. Despite its smaller size and faster inference, EP achieves near-SotA performance in ID testing and significantly better OoD robustness compared to benchmark models. This work represents an initial step toward a trajectory prediction model that is capable of generalizing to different datasets and is robust under changes in sensor setup, scenario selection strategy or post-processing of training
% data.

\section*{ACKNOWLEDGMENT}
Authors would like to thank Shengchao Yan, Wolfram Burgard for contributions to our previous study, and Andreas Philipp for many useful discussions. We also acknowledge
funding from the German Federal Ministry for Economic Affairs and Climate Action within the project ``NXT GEN AI METHODS - Generative Methoden fuer Perzeption, Praediktion und Planung''.

\bibliography{UNIDAS}

\begin{thebibliography}{10}

\bibitem{wilson_argoverse2_2021}
B.~Wilson, W.~Qi, T.~Agarwal, J.~Lambert, J.~Singh, S.~Khandelwal, B.~Pan, R.~Kumar, A.~Hartnett, J.~K. Pontes, {\em et~al.}, ``Argoverse 2: Next generation datasets for self-driving perception and forecasting,'' in {\em Thirty-fifth Conference on Neural Information Processing Systems Datasets and Benchmarks Track (Round 2)}, 2021.

\bibitem{ettinger_waymo_2021}
S.~Ettinger, S.~Cheng, B.~Caine, C.~Liu, H.~Zhao, S.~Pradhan, Y.~Chai, B.~Sapp, C.~R. Qi, Y.~Zhou, {\em et~al.}, ``Large scale interactive motion forecasting for autonomous driving: The waymo open motion dataset,'' in {\em Proceedings of the IEEE/CVF International Conference on Computer Vision}, pp.~9710--9719, 2021.

\bibitem{yao_improving_2024}
Y.~Yao, S.~Yan, D.~Goehring, W.~Burgard, and J.~Reichardt, ``Improving out-of-distribution generalization of trajectory prediction for autonomous driving via polynomial representations,'' in {\em 2024 IEEE/RSJ International Conference on Intelligent Robots and Systems (IROS)}, pp.~488--495, IEEE, 2024.

\bibitem{hestness_deep_2017}
J.~Hestness, S.~Narang, N.~Ardalani, G.~Diamos, H.~Jun, H.~Kianinejad, M.~M.~A. Patwary, Y.~Yang, and Y.~Zhou, ``Deep learning scaling is predictable, empirically,'' {\em CoRR}, vol.~abs/1712.00409, 2017.

\bibitem{cheng_forecast_2023}
J.~Cheng, X.~Mei, and M.~Liu, ``Forecast-mae: Self-supervised pre-training for motion forecasting with masked autoencoders,'' in {\em Proceedings of the IEEE/CVF International Conference on Computer Vision}, pp.~8679--8689, 2023.

\bibitem{zhou_query_2023}
Z.~Zhou, J.~Wang, Y.-H. Li, and Y.-K. Huang, ``Query-centric trajectory prediction,'' in {\em Proceedings of the IEEE/CVF Conference on Computer Vision and Pattern Recognition}, pp.~17863--17873, 2023.

\bibitem{buhet_plop_2020}
T.~Buhet, E.~Wirbel, A.~Bursuc, and X.~Perrotton, ``Plop: Probabilistic polynomial objects trajectory planning for autonomous driving,'' {\em CoRR}, vol.~abs/2003.08744, 2020.

\bibitem{su_temporally_2021}
Z.~Su, C.~Wang, H.~Cui, N.~Djuric, C.~Vallespi-Gonzalez, and D.~Bradley, ``Temporally-continuous probabilistic prediction using polynomial trajectory parameterization,'' in {\em 2021 IEEE/RSJ International Conference on Intelligent Robots and Systems (IROS)}, pp.~3837--3843, IEEE, 2021.

\bibitem{reichardt_trajectories_2022}
J.~Reichardt, ``Trajectories as markov-states for long term traffic scene prediction,'' in {\em 14-th UniDAS FAS-Workshop}, p.~14, 2022.

\bibitem{feng_unitraj_2025}
L.~Feng, M.~Bahari, K.~M.~B. Amor, {\'E}.~Zablocki, M.~Cord, and A.~Alahi, ``Unitraj: A unified framework for scalable vehicle trajectory prediction,'' in {\em European Conference on Computer Vision}, pp.~106--123, Springer, 2025.

\bibitem{yao_empirical_2023}
Y.~Yao, D.~Goehring, and J.~Reichardt, ``An empirical bayes analysis of object trajectory representation models,'' in {\em IEEE 26th International Conference on Intelligent Transportation Systems (ITSC)}, pp.~902--909, IEEE, 2023.

\end{thebibliography}
\bibliographystyle{ieeetr}

%\newpage
\begin{appendices}
\section*{Appendix A: Complexity of Prediction Task}
\label{secs: analyse prediction complexity}
We analyze the complexity of the prediction task in the Waymo Motion dataset under varying history lengths, corresponding to different start times $t_{start} \in [1.1s, 5s]$. Note that $t_{start}=1.1s$ corresponds to the original setting of the Waymo Motion dataset, while 
$t_{start}=5s$ reflects the new setting introduced by our homogenization protocol. 

The complexity is evaluated by quantifying how much the agent's motion deviates from a constant velocity model. Specifically, we measure the longitudinal and lateral deviations between the start position $\mathbf{p}_{start}$ and end position $\mathbf{p}_{end}$ of the ground truth trajectory, normalized by the constant velocity model over a 4.1-second prediction horizon:
%with a 4.1-second prediction horizon $t_{pred}$, where $t_{pred} = t_{end} - t_{start}$ :
\begin{equation}
\begin{aligned}
\label{eqn:normlized distance}
\mathbf{d} &= \frac{(\mathbf{p}_{end}-\mathbf{p}_{start})\boldsymbol{R}_{start}}{|\mathbf{v}_{start}|* 4.1s}\\
\end{aligned}
\end{equation}

\noindent where $\boldsymbol{R}_{start}$ is the $2\times 2$ rotation matrix determined by the agent's heading at $t_{start}$ and $|\mathbf{v}_{start}|$ denotes the velocity magnitude at time ${t_{start}}$. $\mathbf{d}$ is a 2D vector of normalized longitudinal and lateral distances. A value of $\mathbf{d}= [1,0]$ indicates that agent's motion strictly follows the constant velocity model. In contrast, larger deviations from $[1,0]$ reflect greater divergence from constant velocity, signifying higher prediction complexity.

\begin{figure}[!th]
%\vspace{+5pt}
\centering
\includegraphics[width=3.8in]{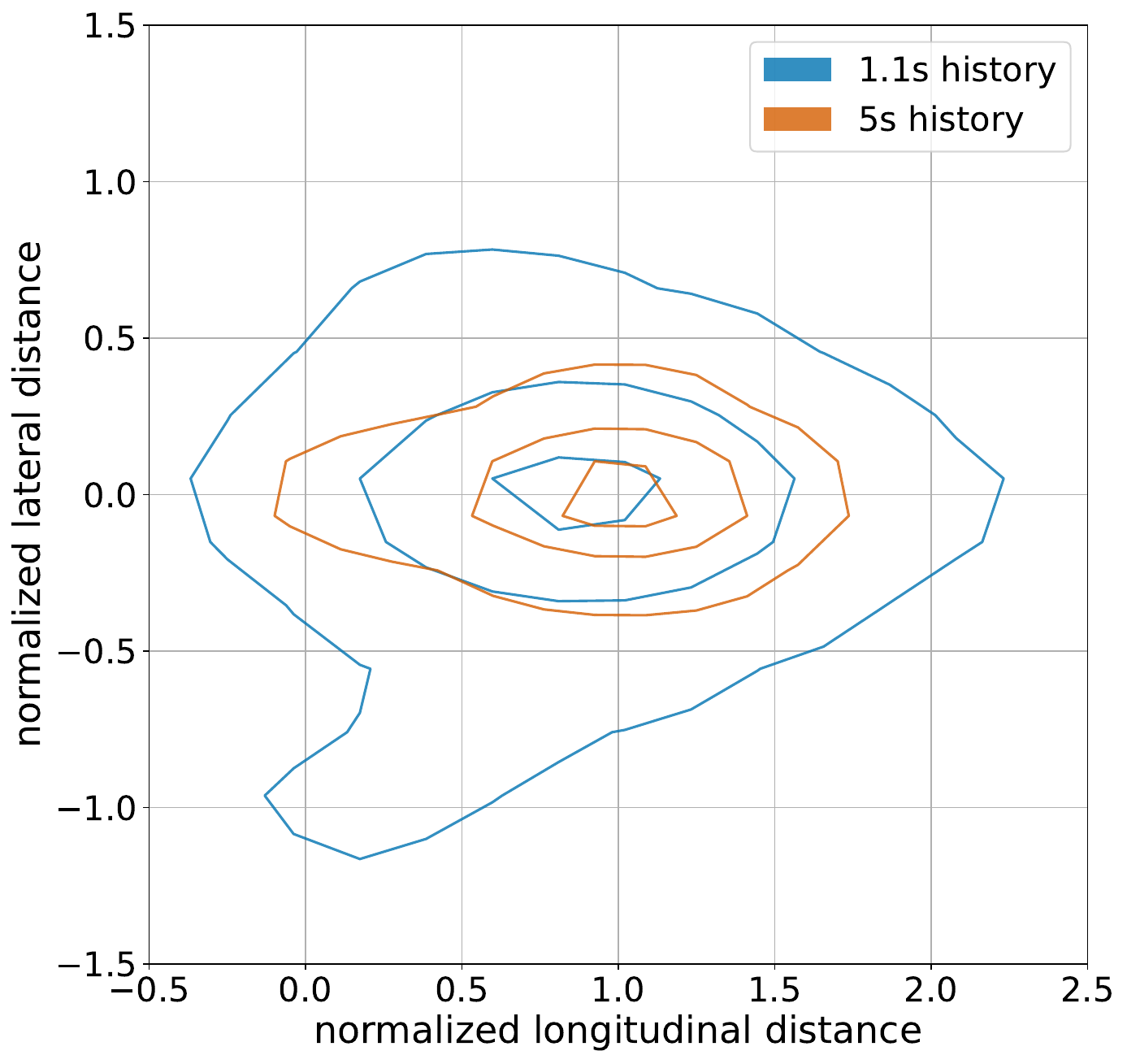}
\caption{Kernel density plot of normalized longitudinal and lateral distances with 1.1-second and 5-second history lengths for Waymo Motion validation set. Contours indicate the 30-th, 60-th and 90-th
percentiles, respectively, illustrating a broader spread for the prediction task with a 1.1-second history.}
\label{fig:prediction_complexity}
\vspace{0.0em}
\end{figure}

We visualize the distribution of $\mathbf{d}$ for $t_{start} \in [1.1s, 5s]$ of Waymo Motion validation set in Figure \ref{fig:prediction_complexity}. Compared to the distribution of $\mathbf{d}$ at $t_{start}=5s$, the distribution at $t_{start}=1.1s$ is broader, indicating that the prediction task with a 1.1-second history is more complex than with a 5-second history.

\end{appendices}

\end{document}